\documentclass{article}
\usepackage{graphicx}
\usepackage{xcolor}
\usepackage[margin=1in]{geometry}
\usepackage{enumitem}
\usepackage[frozencache]{minted}

\newcommand{\figscale}{0.8}

\title{Project Florida: Federated Learning Made Easy}
\author{Daniel Madrigal Diaz, Andre Manoel, Jialei Chen, Nalin Singal, Robert Sim \\ \{danielmad,andre.manoel,jialeichen,rsim\}@microsoft.com, singal.nalin@gmail.com
\\[2ex] Microsoft Corporation}
\date{\today}

\begin{document}

\maketitle

\begin{abstract}
    We present \emph{Project Florida}, a system architecture and software development kit (SDK) enabling deployment of large-scale Federated Learning (FL) solutions across a heterogeneous device ecosystem.  Federated learning is an approach to machine learning based on a strong data sovereignty principle- i.e.\ that privacy and security of data is best enabled by storing it at its origin, whether on end-user devices or in segregated cloud storage silos.  Federated learning enables model training across devices and silos while the training data remains within its security boundary, by distributing a model snapshot to a client running inside the boundary, running client code to update the model, and then aggregating updated snapshots across many clients in a central orchestrator. Deploying a FL solution requires implementation of complex privacy and security mechanisms as well as scalable orchestration infrastructure.  Scale and performance is a paramount concern, as the model training process benefits from full participation of many client devices, which may have a wide variety of performance characteristics.  Project Florida aims to simplify the task of deploying cross-device FL solutions by providing cloud-hosted infrastructure and accompanying task management interfaces, as well as a multi-platform SDK supporting most major programming languages including C++, Java, and Python, enabling FL training across a wide range of operating system (OS) and hardware specifications. The architecture decouples service management from the FL workflow, enabling a cloud service provider to deliver FL-as-a-service (FLaaS) to ML engineers and application developers.  We present an overview of Florida, including a description of the architecture, sample code, and illustrative experiments demonstrating system capabilities.
\end{abstract}

\section{Introduction}

Federated learning~\cite{mcmahan2017communication, https://doi.org/10.48550/arxiv.1812.06127, https://doi.org/10.48550/arxiv.1910.06378} is a machine learning paradigm that enables multiple entities to collaboratively train a shared model without sharing their own data. This approach can protect the privacy and security of the data owners, and leverage the diversity and richness of distributed data sources. For example, federated learning can be used to train a speech recognition model across different devices, or a text prediction model across different end user smartphones~\cite{hosseini2021federated,stremmel2020pretraining} . Each entity/client locally trains a model on its own data, and periodically communicates with a central server that coordinates the model updates and aggregates them into a global model. This approach can offer benefits such as privacy preservation, bandwidth reduction, and scalability. However, implementing a platform to do federated learning also poses several challenges, such as:
\begin{itemize}
\item \textbf{Communication efficiency:} Federated learning requires frequent and reliable communication between the central server and the participating devices or nodes, which can incur high bandwidth and latency costs, especially for large-scale and mobile scenarios. The platform needs to optimize the communication protocols, compression techniques, and synchronization strategies to reduce the communication overhead and ensure the quality of service.
\item \textbf{Client heterogeneity:} The clients may have different hardware, software, network, and power capabilities, which can affect the efficiency and reliability of the federated learning process. For example, some clients may have faster or more powerful devices than others, some clients may have different software versions or platforms, some clients may have unstable or limited network connections, and some clients may have low battery or high mobility. These factors can introduce latency, communication overhead, or failure in the federated learning process, and require careful design of the federated learning protocol, such as compression, synchronization, or fault-tolerant methods.
\item \textbf{Privacy and security:} Federated learning aims to protect the data privacy and security of the participants, but it also faces potential threats from malicious or compromised parties, such as data leakage, model poisoning and inference attacks. The platform needs to incorporate various privacy and security mechanisms, such as encryption, differential privacy, secure aggregation, and client attestation, to prevent or mitigate these attacks and ensure the trustworthiness of the federated learning process.
\item \textbf{Asynchrony:} FL workflows benefit from increased scale, which can lead to slow convergence when the training workflow is synchronously orchestrated.  Asynchronous protocols such as Papaya~\cite{DBLP:journals/corr/abs-2111-04877}, which enable the orchestrator to merge late-arriving results can enable scaling to millions or hundreds of millions of connected devices.
\end{itemize}

To address these challenges we developed Project Florida, a cross-device federated learning platform that aims to facilitate the development, deployment, and evaluation of federated learning applications in various domains and scenarios. Florida is designed to decouple service deployment and management from the application and machine learning lifecycles, enabling cloud service providers to deliver Federated Learning as a Service (FLaaS) solutions to a variety of customers interested to deploy FL workflows. This decoupling was validated in~\cite{mandke:interspeech23}, where a third-party speech verification system was deployed using a first-party managed Florida deployment for orchestration. 

In the following sections, we will present the Florida architecture, describe each of its components in detail, and demonstrate the use of the platform to train a spam classification model.  The main contributions of this work can be summarized as follows:

\begin{itemize}
\item An overview of Florida, it's architecture and main features.
\item A breakdown of FLaaS tasks relative to application developer, service engineer, and data scientist personas, and how Florida cleanly separates these personas.
\item Experimental validation of the platform, including model training and scale testing
\item A discussion of future opportunities for FLaaS platforms.
\end{itemize}

\section{Florida Platform}
Florida is a cross-device federated learning platform designed to support production-oriented scenarios where data privacy, efficiency and robustness are crucial. Florida provides cross-platform SDKs that enable heterogeneous devices to communicate with the server, train and evaluate models locally, and use cross-compatible cryptographic functions to protect model updates. It also offers back-end services that run on cloud infrastructure that securely orchestrate the devices and aggregate the incoming model updates. Additionally, it has user interfaces (CLI/GUI) that allow users to easily create, manage and monitor federated learning tasks.

Florida provides a FLaaS solution that enables developers and ML researchers to create and deploy applications with federated learning capabilities. It abstracts the complexity of federated learning algorithms, communication protocols, and security mechanisms, and provides user-friendly APIs. Florida's features include:

\begin{itemize}
\item \textbf{ML Framework Agnostic:} Use the training and inference framework best suited to the application. 
\item \textbf{General-purpose:} Support for typical weighted FL aggregation schemes such as FedAvg~\cite{mcmahan2017communication}, FedProx~\cite{DBLP:journals/corr/abs-1812-06127}, and DGA~\cite{DBLP:journals/corr/abs-2106-07578}.
\item \textbf{Built-in Privacy and Security Features:} Differential Privacy and Secure Aggregation are included out-of-the-box.
\item \textbf{Support for Heterogeneous Devices:} A fully cross-platform SDK with mutually compatible security primitives, enabling deployment of FL tasks across a heterogeneous device ecosystem.  Currently supported languages include C\#, Javascript/TypeScript, C++, Kotlin (Java) and Python.
\item \textbf{Support for Synchronous or Asynchronous FL:} a Florida task can be configured for synchronous or asynchronous updates from clients, enabling straggler devices to contribute work. 
\item \textbf{Florida Dashboard and CLI:} a web UI that allows ML scientists to create, manage and monitor FL tasks. The UI also enables visualizing metrics such as accuracy, loss, etc; and a command-line interface for scripting service and workflow management.
\item \textbf{Security:} User access control and security group permissions to enable collaborative FL task management.
\item \textbf{Android Device Attestation:} Support to validate Google Play Integrity verdicts and Huawei SysIntegrity.
\end{itemize}

\subsection{User Personas}

Florida is designed with a set of distinct user personas in mind.  Production FL systems require a variety of development efforts that leverage widely different skill sets.  Furthermore, the economics of building and deploying FL solutions suggests the need for reusable orchestration architecture which can span a range of applications and/or organizations. In principle, a single service deployment could service multiple independent customers with their own application provisioning and ML toolchains.  Table~\ref{tab:personas} describes the three main personas in mind while designing Florida, and illustrates how development tasks are clearly delineated between the three.  Florida is modularized such that service deployment and management can be completely decoupled from the application development life cycle.  Furthermore, the platform minimizes requirements on any single ML framework, enabling the ML engineer to select the framework(s) that are best suited to the device capabilities and learning task.

\begin{table}[]
    \centering
    \begin{tabular}{|p{1.7in}|p{1.7in}|p{1.7in}|}
    \hline
        \textbf{Devops Engineer} & \textbf{Application Engineer} & \textbf{ML Engineer} \\ \hline 
        \begin{itemize}[left=0pt,topsep=0pt]        
        \item Build and manage service container images
        \item Configure compute, scale-out parameters
        \item Deploy services
        \item (optionally) manage metering and billing for 3P access
        \end{itemize}
        & 
        \begin{itemize}[left=0pt,topsep=0pt]
        \item Responsible for application lifecycle (e.g. touch keyboard app)
        \item Manages app deployment to device ecosystem
        \item Defines app-dependent ML use case
        \item Sets task-, app-, or device-dependent criteria for triggering local training
        \item coordinates with ML engineer to define ML and data lifecycle components in the context of the app.
        \end{itemize}
        & 
        \begin{itemize}[left=0pt,topsep=0pt]
        \item Develops on-device data pipeline, training loop
        \item Integrates training loop with Florida SDK
        \item Coordinates with application developer to deploy training code to devices via standard application deployment processes
        \item Configures and manages FL tasks in Florida service
        \end{itemize} \\ \hline
    \end{tabular}
    \caption{FL activities by persona.  Florida enables strict decoupling between roles such that, in particular, the devops role can be organizationally separated from that of the application and ML engineers.}
    \label{tab:personas}
\end{table}

\section{Florida Architecture}

\begin{figure}
    \centering
    \includegraphics[width=\figscale\textwidth]{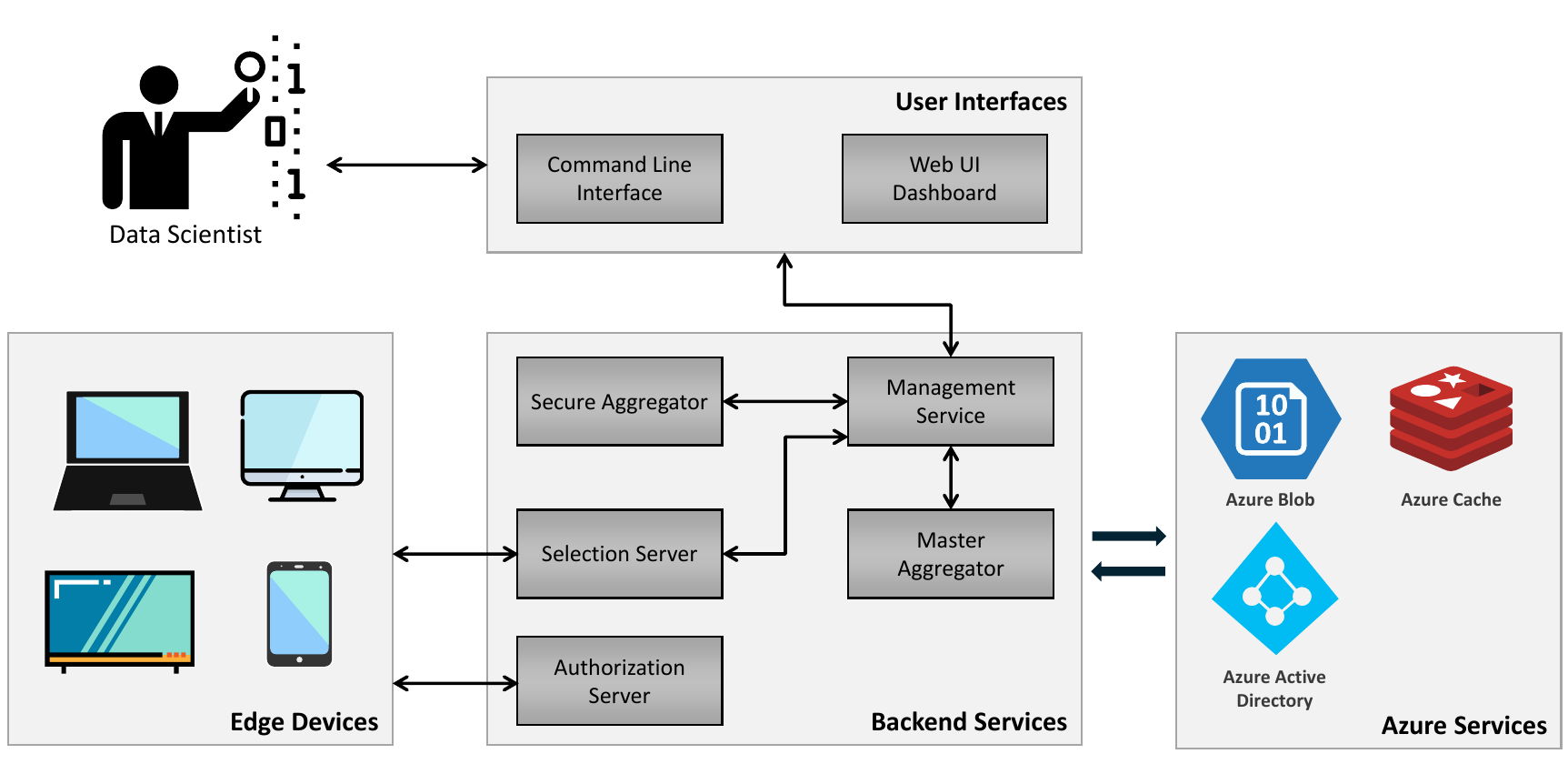}
    \caption{Florida architecture.}
    \label{fig:architecture}
\end{figure}

Florida's design involves two high-level components (Figure \ref{fig:architecture}): a collection of cloud-based back-end services that provide orchestration, data processing and storage, and a local client service that runs on connected edge devices, such as smartphones, laptops, tablets, etc.  Florida services are implemented in ASP.Net and containerized for deployment to scalable Kubernetes clusters. Task state is managed using a Redis cache.

\subsection{Back-end Services}

\subsubsection{Management Service}
The main component of Florida is the \textit{Management Service}, which has two major roles:
\begin{itemize}
\item \textbf{User Interface API:} receive requests from administrator-facing interfaces to create, manage, and monitor federated learning tasks.
\item \textbf{Task Orchestrator:} advertising the available tasks to the \textit{Selection Service}, sending requests to the Aggregator services to perform secure aggregation and central model update, and monitoring overall task progress. 
\end{itemize}

\subsubsection{Secure Aggregator}
 Florida has a two-stage aggregation process similar to~\cite{bonawitz2019towards}. In the first stage, the \textit{Secure Aggregator} groups clients into subsets called Virtual Groups (VGs). Then it sends instructions to the clients in each VG to negotiate secure aggregation steps using a secure multi-party computation (MPC) protocol~\cite{DBLP:journals/corr/BonawitzIKMMPRS16}. Finally, it receives the local updates from the clients and aggregates them, creating an interim result for each VG. (Section \ref{sec:sec_aggregation}). The purpose of aggregating in VGs is to address the performance cost of the secure MPC protocol, which scales with $O(n^2)$, where $n$ is the number of participating clients in a VG. VGs should be large enough to provide reasonable security and privacy guarantees while managing the quadratic cost of running the secure protocol.

\subsubsection{Master Aggregator}
The second aggregation stage involves the \textit{Master Aggregator}. This component aggregates all the interim results that the Secure Aggregators generate in the first stage, where they aggregate the local updates from the clients. The Master Aggregator then updates the global model with the aggregated results, using the user-defined logic. The user-defined logic can be specified as python script, a Linux native executable file, or as an Azure ML pipeline API.

\begin{figure}
    \centering
    \includegraphics[width=\figscale\textwidth]{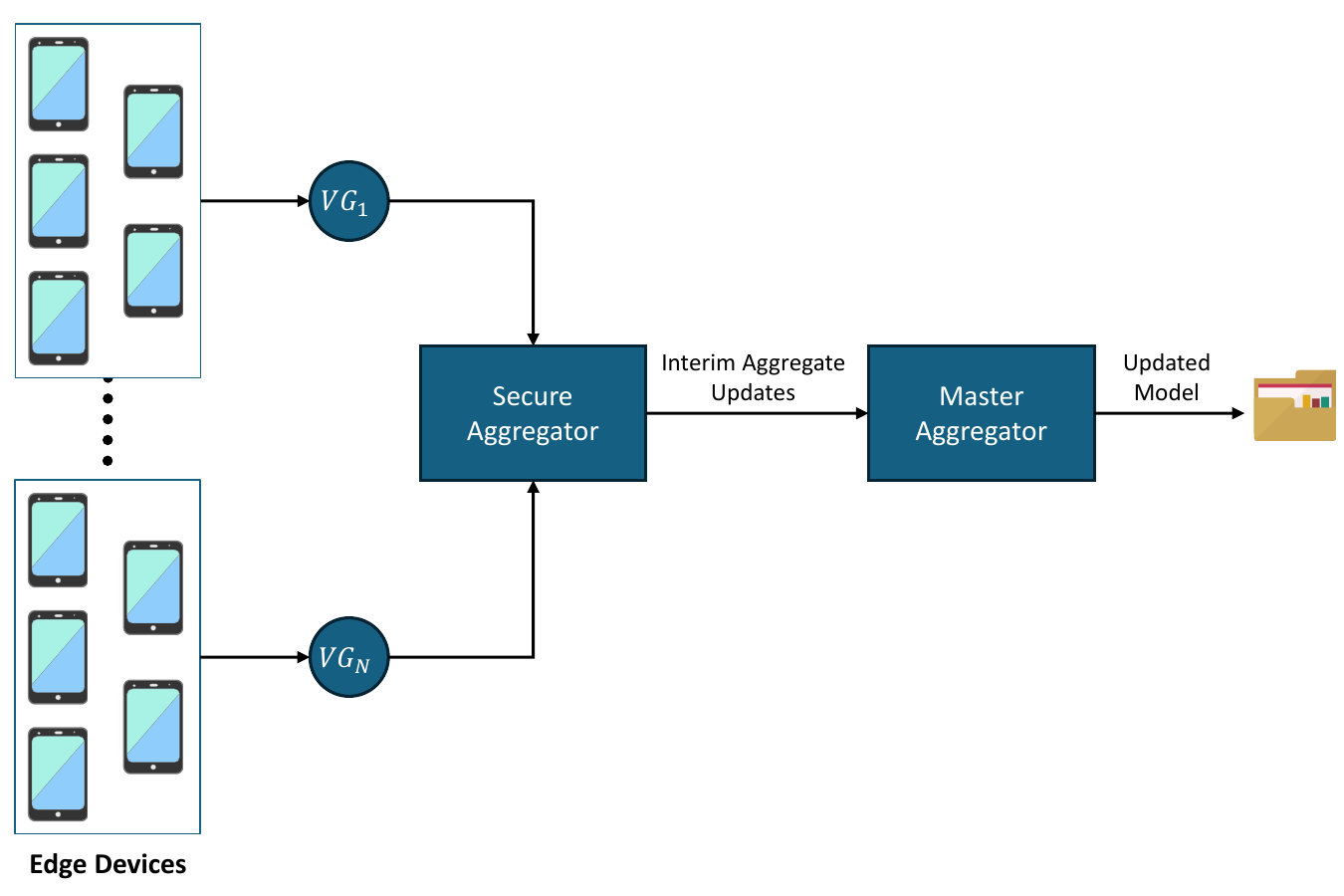}
    \caption{Two-stage aggregation process.}
    \label{fig:aggregation}
\end{figure}

\subsubsection{Selection Service}
The \textit{Selection Service}  facilitates communication between clients and the server. It acts as an intermediary between the two, advertising available tasks to clients and registering them as participants when they meet the requirements. Once enough clients have registered, the Selection Service randomly selects a subset of participants and provides them with the task details. This ensures that the workload is distributed evenly among participants. Additionally, the Selection Service tracks the training status of the participants and provides additional instructions when necessary. It is responsible for ensuring that clients are matched with appropriate tasks that they can complete successfully, 

\subsubsection{Authentication Service}
The \textit{Authentication Service} protects the system from malicious or compromised devices. Before a device can join a federated learning task, it must provide an attestation certificate to prove its integrity. This certificate contains information about the device's hardware, software, and configuration, and is verified by a trusted third-party service. The Authentication Service is responsible for validating the attestation certificate and determines whether the device fulfills the criteria. This ensures that only devices that meet the required integrity standards are allowed to participate in the federated learning job. Currently, the Authentication Service provides out-of-the-box mechanisms to validate Google Play Integrity verdicts and Huawei SysIntegrity Responses~\cite{goog:integrity, huaw:integrity}.

\subsection{Client SDK}

The Florida SDK enables application developers to easily integrate their client-side training workflow with Florida.  Language bindings are currently available for Python, Kotlin/Java, .NET, C++, and Javascript. Clients have the option to communicate with the Florida service using gRPC\cite{grpc} or REST.  Crucially, the various client libraries employ mutually compatible cryptographic key derivation primitives, so that, for example, secret key negotiation for secure aggregation is successful across different languages and operating systems. 

\begin{figure}
    \centering
    \includegraphics[width=\figscale\textwidth]{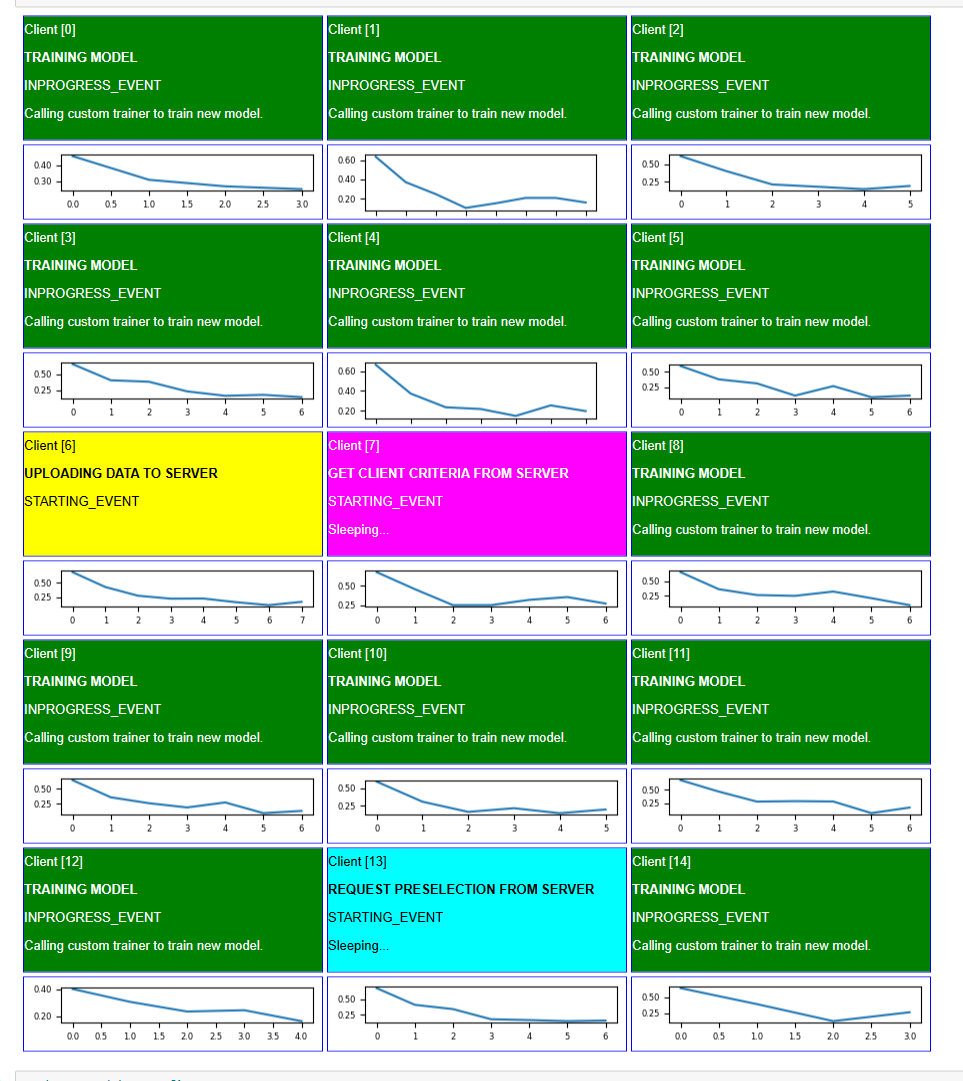}
    \caption{Jupyter noteboook example running 15 pytorch clients on a spam training task.}
    \label{fig:notebook}
\end{figure}

Figure~\ref{fig:sdk_python} illustrates a simple client workflow implementation in python. The application developer can substitute the trainer for a full training loop, for example using scikit-learn~\cite{scikit-learn}, PyTorch~\cite{NEURIPS2019_9015}, ONNX~\cite{onnx}, or Tensorflow~\cite{Abadi+16}.  Figure~\ref{fig:notebook} illustrates a toy example of 15 clients training in a Jupyter notebook. Individual client states are represented in each pane, and updated interactively using iPyWidgets.

\begin{figure}
    \centering
    \begin{minipage}{0.85\textwidth}
    \input{figures/code_snippet.tex}
    \end{minipage}
    \caption{Sample Python client.}
    \label{fig:sdk_python}
\end{figure}

\subsection{User Interfaces}

Florida provides a web-based GUI or command line interface (CLI) for FL task management.  The CLI enables the same functionality provided by the GUI while enabling task configuration through secured deployment environments. Below we summarize the features provided by the web UI.

\subsubsection{Florida Dashboard (WebUI)}

The Florida dashboard enables authenticated users to create and manage training workflows. 

\begin{figure}[htb]
    \centering
    \includegraphics[width=\figscale\textwidth]{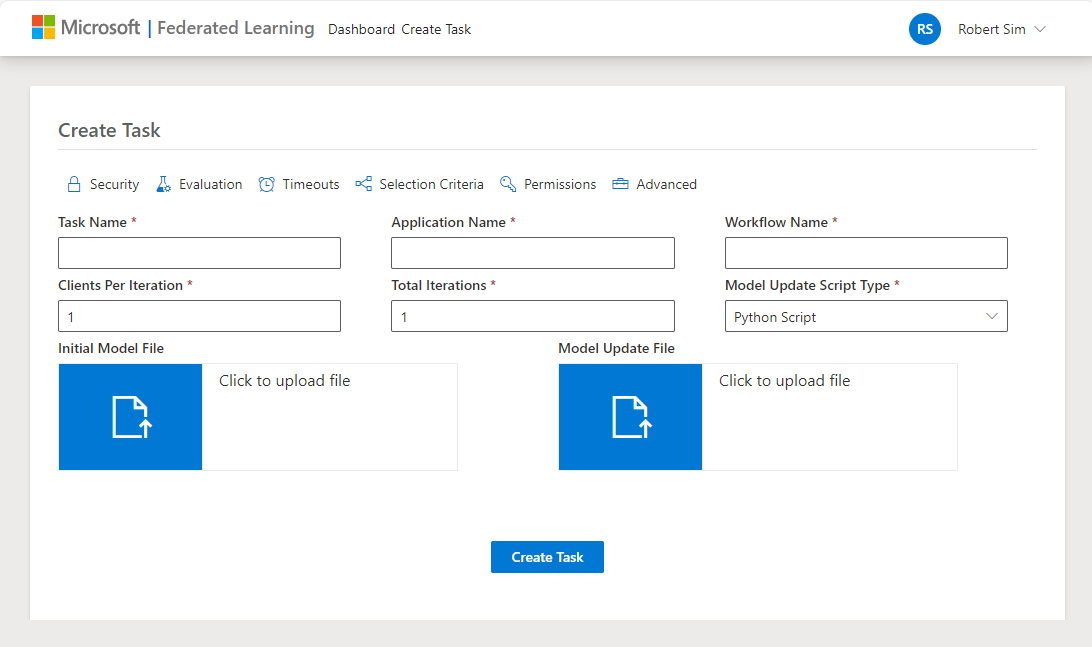}
    \caption{Task creation interface.}
    \label{fig:taskcreate}
\end{figure}

\paragraph{Task Creation:} On creating a new task (Figure~\ref{fig:taskcreate}), the ML scientist specifies a task name, application name, and workflow as defined below, as well as the number of desired clients per round and total number of rounds to execute the workflow.  The operator also uploads an initial model snapshot, and provides a python script or .NET executable for performing master aggregation. The operator can optionally also specify the security and privacy configuration, set selection criteria for device participation, and set task permissions to enable sharing with other ML scientists.  

The text fields defining a task are defined as follows:
\begin{itemize}
\item \textbf{Task name}: This name defines the current FL task and enables the ML scientist to differentiate between different configured tasks.
\item \textbf{Application name}: This name is application specific and used by connecting devices to associate the particular on-device application with available workflows.
\item \textbf{Workflow name}: This name defines a particular workflow on the connected device.  For instance a deployed application may have training workflows for next word prediction or preference ranking. This name enables the client-side application to choose the appropriate workflow for the task. 
\end{itemize}

\begin{figure}[htb]
    \centering
    \includegraphics[width=\figscale\textwidth]{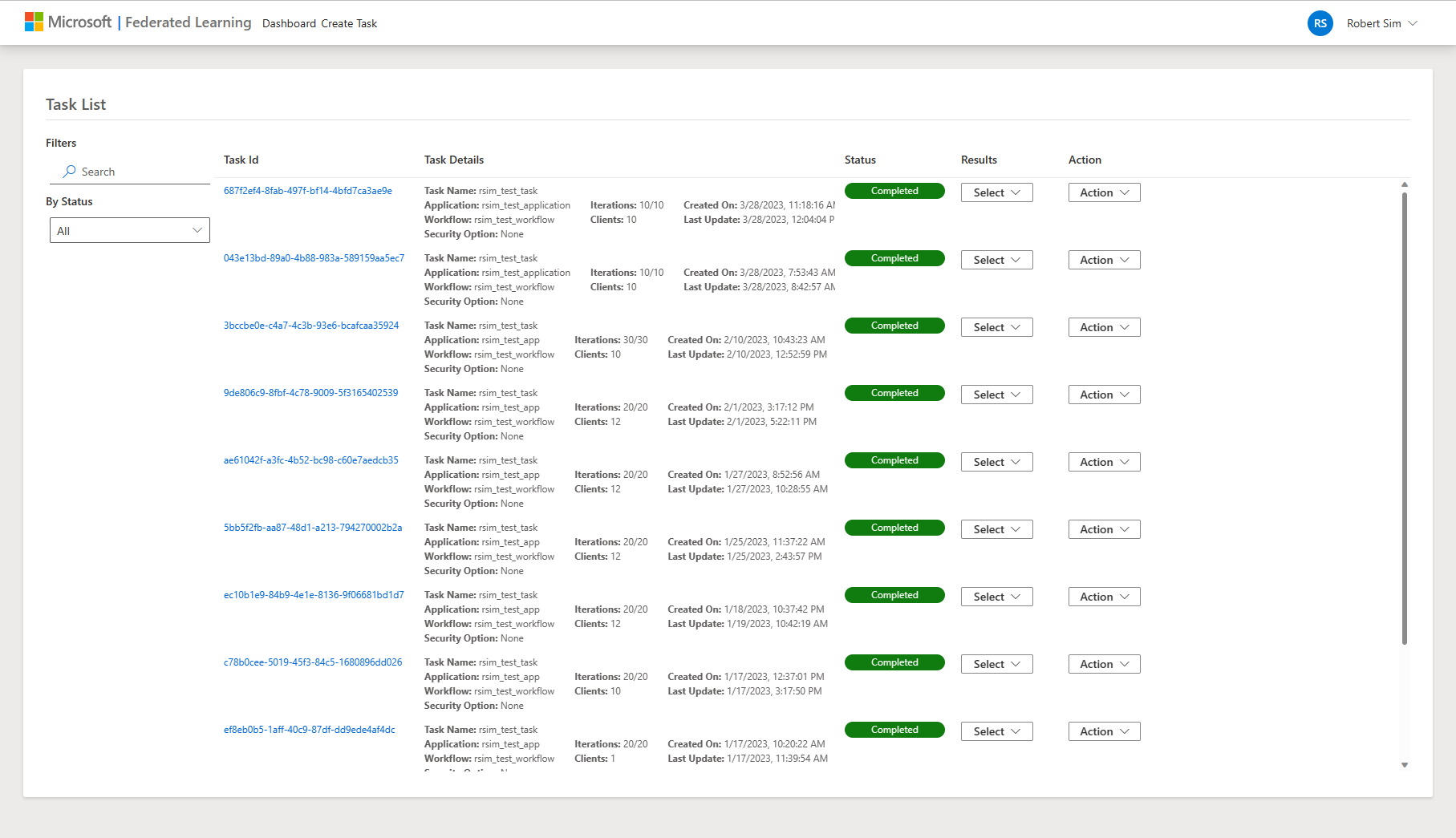}
    \caption{Task management interface.}
    \label{fig:taskmanagement}
\end{figure}

\paragraph{Task Management:}  The task management interface (Figure~\ref{fig:taskmanagement}) is the default page and enables the operator to see at a glance the configured tasks and their state (running, paused, completed, etc).  A brief summary of task properties is includes, such as number of rounds and connected devices.  Clicking on any task launches the task view.

\begin{figure}
    \centering
    \includegraphics[width=\figscale\textwidth]{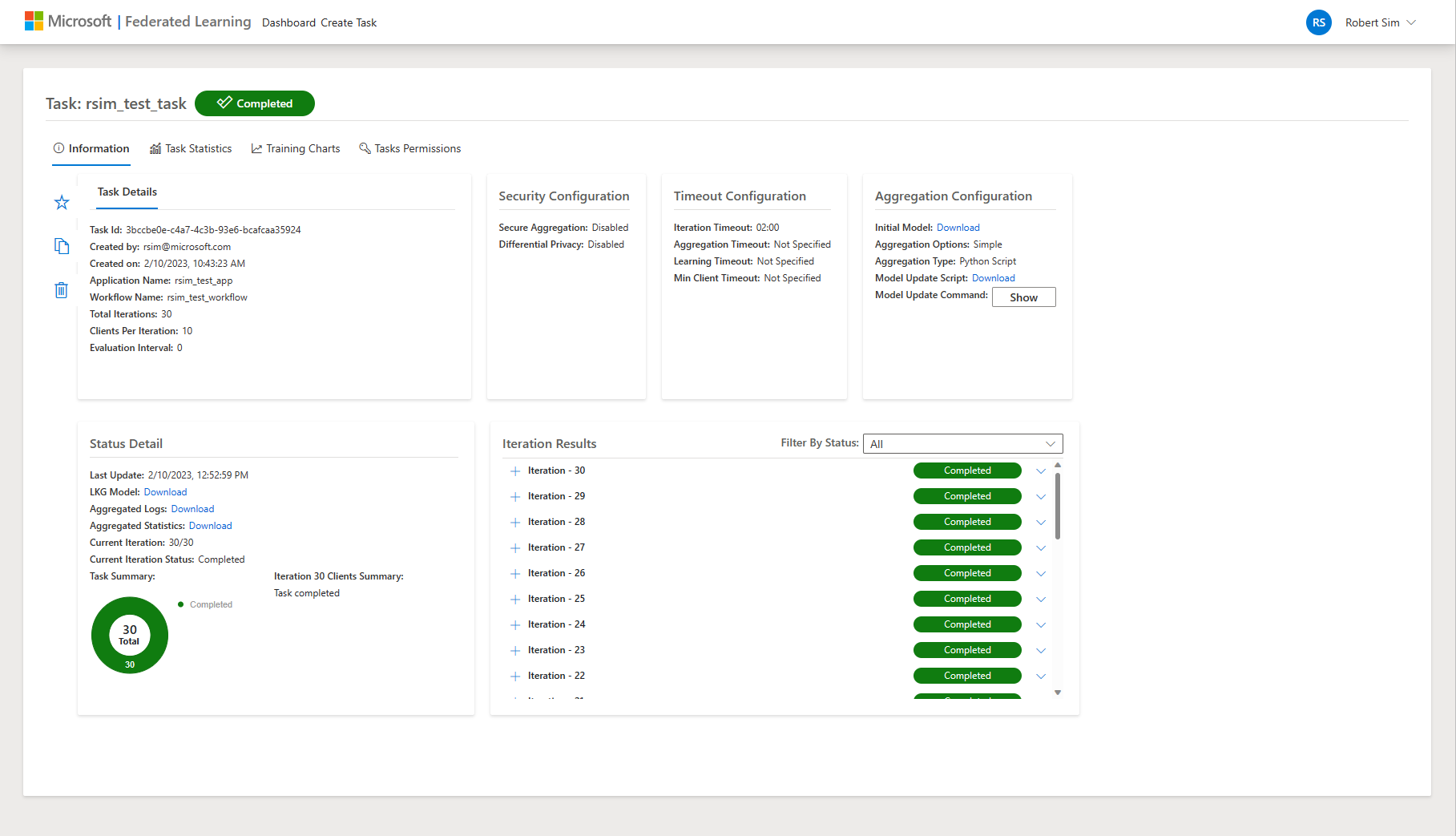}
    \caption{Task view.}
    \label{fig:taskview}
\end{figure}

\paragraph{Task View:}  The task view (Figure~\ref{fig:taskview}) displays a selected task and its state.  Results from any training round can be accessed, as well as drill-down to client- or server-side metrics describing convergence of the training process (the loss), model performance, and run-time performance of the training task.  The task view also enables the operator to pause or cancel a running task.

\begin{figure}
    \centering
    \includegraphics[width=\figscale\textwidth]{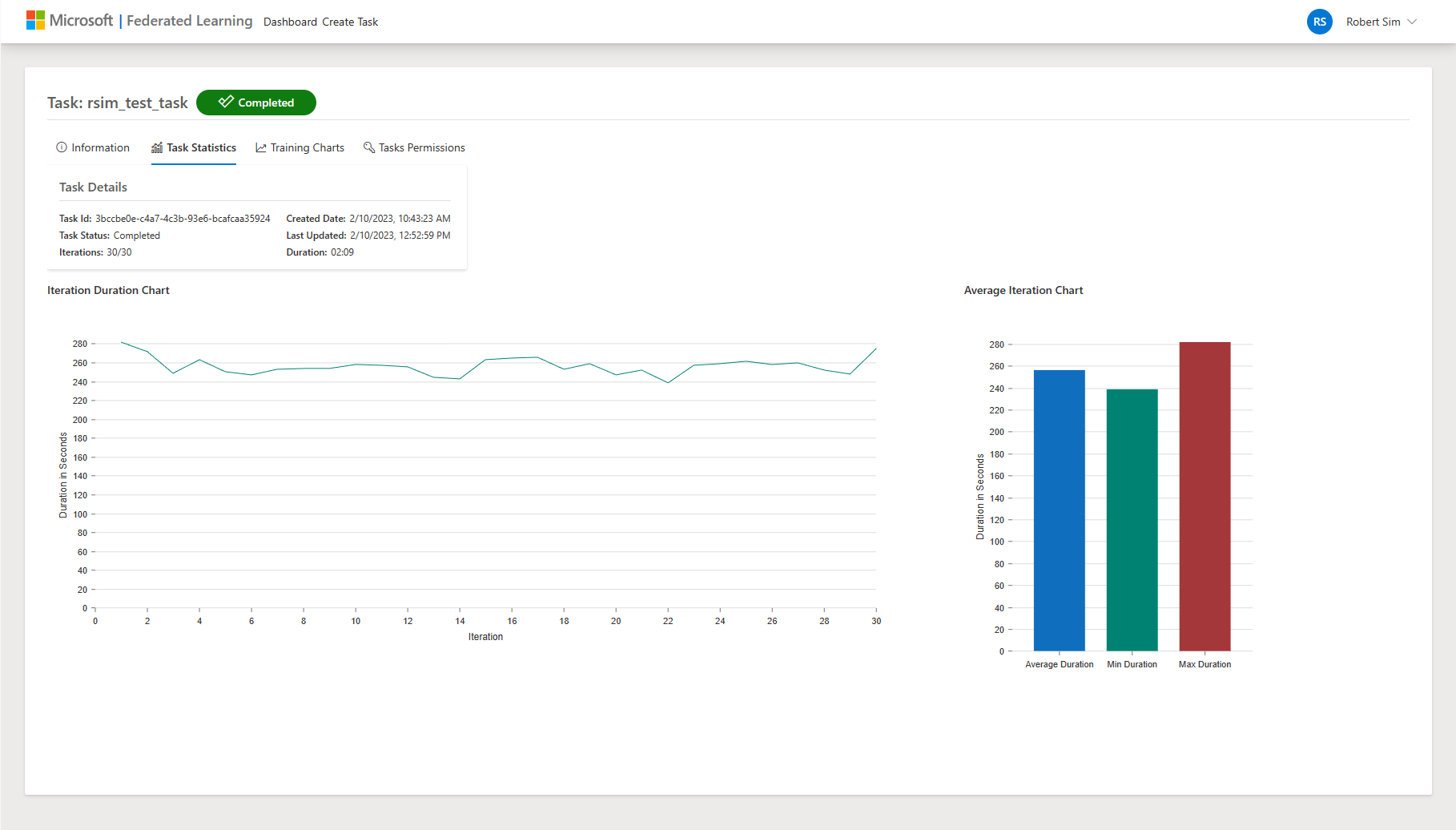}
    \caption{Training performance statistics.}
    \label{fig:statistics}
\end{figure}

\begin{figure}
    \centering
    \includegraphics[width=\figscale\textwidth]{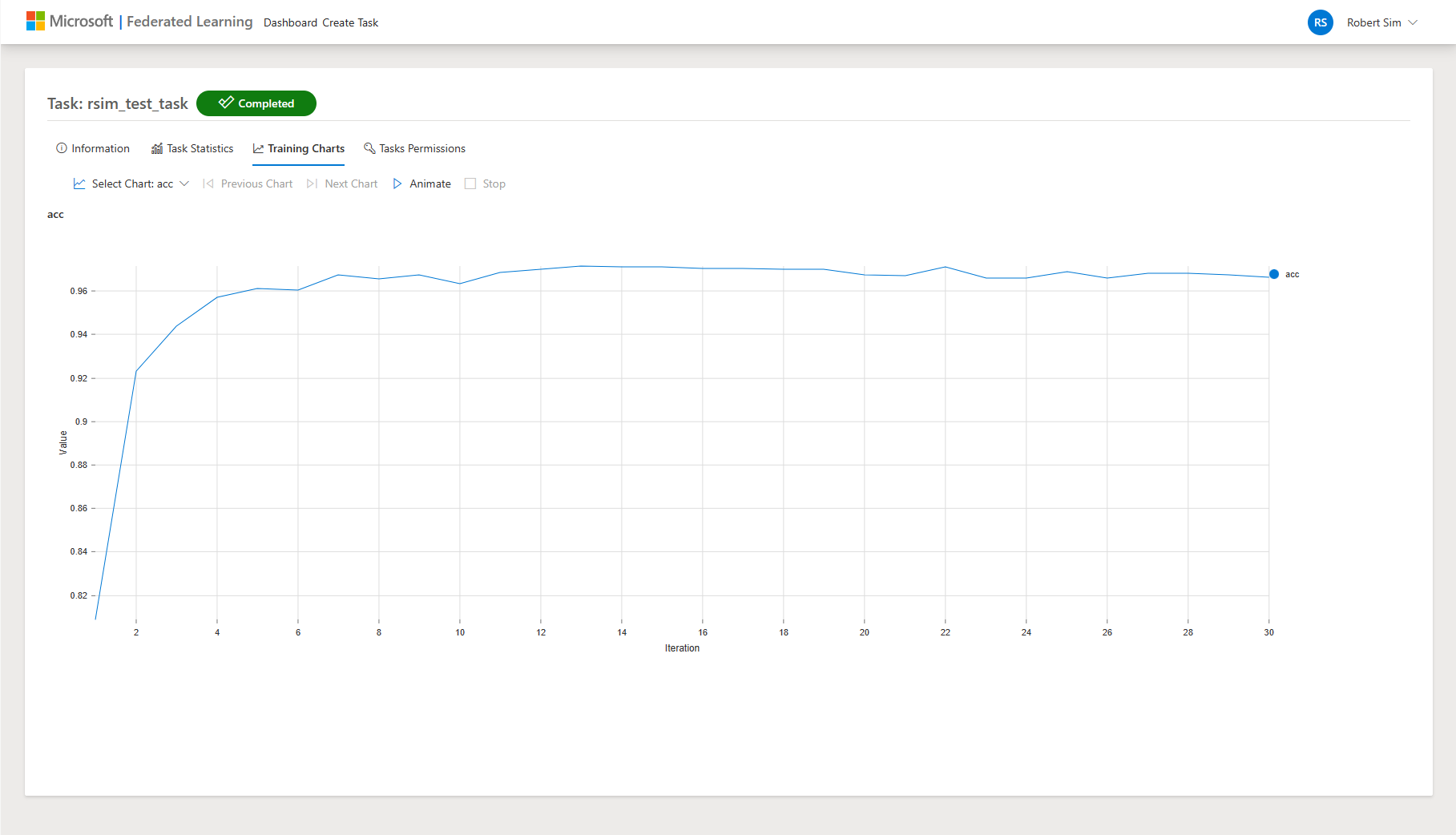}
    \caption{Task evaluation metrics.}
    \label{fig:evaluationmetrics}
\end{figure}

\paragraph{Metrics:} Drilling down from the task view, the ML scientist can view training and run-time performance metrics (Figure~\ref{fig:statistics}), as well as evaluation metrics such as model accuracy (Figure~\ref{fig:evaluationmetrics}).  The task configuration can specify the interval at which to evaluate client- or server-side evaluation data.

\section{Security and Privacy}

\subsection{Secure Aggregation}
\label{sec:sec_aggregation}

The security of federated learning depends on the integrity of the central orchestrator.  Most FL schemes assume an ``honest-but-curious'' orchestrator, which implements the orchestration protocol honestly, but may attempt to recover private information from the uploaded model weights. One solution to securing the uploaded data is to apply a secure multi-party computation approach known as \textit{secure aggregation (SA)}~\cite{DBLP:journals/corr/BonawitzIKMMPRS16}. SA operates by exchanging symmetric key pairs between pairs of clients using Diffie-Hellman key exchange, and generating a mask between each pair such that the sum of the masked model weights or pseudo-gradients from each client is equal to the sum of the model weights/pseudo-gradients itself. That is, to compute the sum $\sum_i x_i$ of model weights $x_i$ from each client $i$, transmit and compute instead the sum of masked payloads $y_i$:
\[
y_i = x_i + \sum_{i<v} s_{i,v} - \sum_{i>v} s_{i,v}
\]
where $s_{u,v}$ is a randomly generated mask of the same dimension as $x_i$ which is shared between clients $u$ and $v$. Thus it can be shown that through cancellation
\[
\sum y_i == \sum x_i 
\]

For secure aggregation to provide strong security it is important that pairs of clients generate cryptographically strong masks, which are applied using modular integer arithmetic. There are two consequences to this requirement: first the model must be quantized and transformed into an array of integers, an operation which can be only partially reversed after the weights are aggregated, and second we face the engineering challenge of generating random masks across heterogeneous operating system platforms. Note that for efficiency client pairs only negotiate a shared secret, and must compute the mask locally. Florida utilizes strong and cross-platform compatible key derivation functions (KDFs)~\cite{wiki:kdf} to ensure consistent mask generation even across different device operating systems.

\subsection{Differential Privacy}

Differential privacy injects Gaussian noise into the training process to ensure the model is approximately invariant to the contributions of any single device~\cite{Abadi_2016}. We provide support for local or global differentially-private noise addition.  On task configuration, the user specifies the local or global mechanism and the desired noise multiplier $\mu$. When DP is enabled, the user can access a R{\'{e}}nyi-DP privacy accountant~\cite{DBLP:journals/corr/abs-1808-00087} in the dashboard to determine the current privacy loss $\epsilon$.

\subsection{Secure Computing and Asynchronous FL}

Confidential containers~\cite{confidential_containers} enable developers to run secure workloads in untrusted cloud environments, and in the case of Federated Learning this can help mitigate the problem of relying on an ``honest-but-curious'' cloud provider. Furthermore, attested confidential containers can yield a much simplified approach to secure aggregation-- provided that the container is vetted and trusted, clients can encrypt and upload model updates without negotiating paired keys with other clients in a virtual group~\cite{DBLP:journals/corr/abs-2111-04877}.  This approach enables fast, asynchronous federation schemes.  Florida supports asynchronous training and may be deployed using confidential containers to ensure the integrity of the aggregation step.

\section{Experimental Results}

\begin{figure}[htp]
    \centering
    \includegraphics[width=0.6\textwidth]{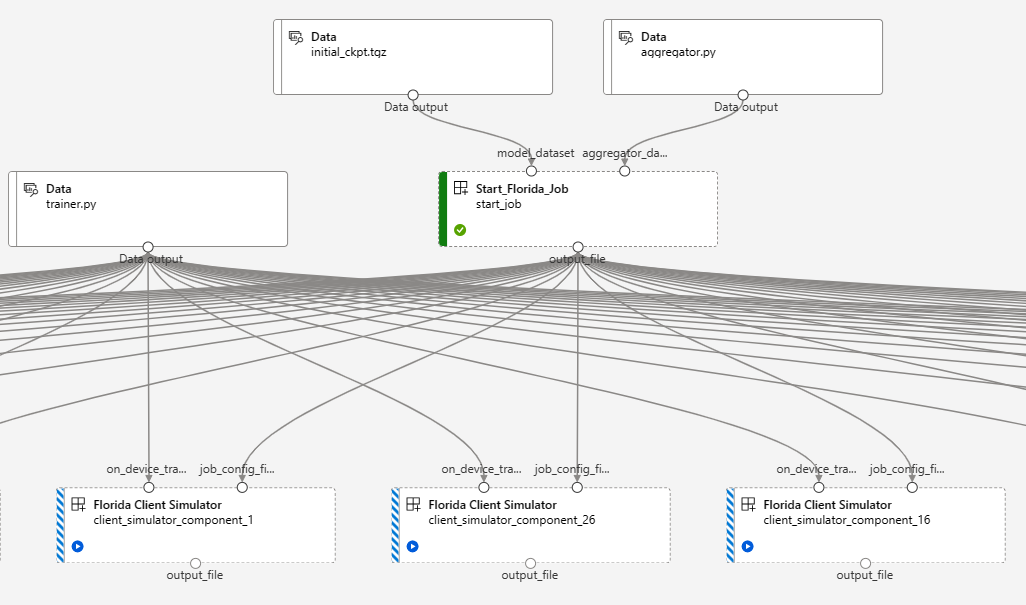}
    \caption{End-to-end simulator on an Azure ML pipeline. The \emph{Start Florida Job} component communicates with the server to create the task on the back end, while each of the \emph{Florida Client Simulator} components have multiple clients running in parallel to execute the task. Finally, a \emph{Run Florida Job} component (not in display above) is responsible for communicating with the server to fetch training status and metrics.}
    \label{fig:aml_simulator}
\end{figure}

We now show how Florida can be used in practice for specific tasks. 
For these experiments, we have the Florida back end services deployed on Azure, and are able to run the clients in a variety of basic test configurations. One possibility is to just run a few clients locally on a single machine, perhaps for an initial test. For this purpose, we have written a Jupyter notebook with widgets that make it easier to follow the training in each client (Figure~\ref{fig:notebook}). For running more than just a few clients, one might want to use our AzureML simulator, which has clients running inside multiple components of an AzureML pipeline (Figure \ref{fig:aml_simulator}), while constantly communicating to the Florida server to fetch training status and metrics, which are communicated via a dashboard on AzureML.

\subsection{Spam Classification}

Consider the (somewhat classic) problem of determining whether a given e-mail or text message is spam; this is usually done by training a binary classification model over a large database, which combines all users data. In some situations, however, it might be desirable to not use such a database, and we would instead train this model locally for each user. While a single user's data might not be enough to provide a good performance, FL allows us to merge model updates from multiple users to build a global model.

In this specific example, we use the Python SDK together with the HuggingFace transformers library\cite{wolf2019huggingface}. The model and dataset come from the HuggingFace Hub -- the model is BERT Tiny\cite{turc2019well} (\texttt{prajjwal1/bert-tiny}), a small transformer obtained by distilling BERT, and the dataset Enron Spam (\texttt{SetFit/enron-spam}), a curated version of the e-mails released during an investigation. For simplicity, we split the dataset in 100 subsets of same size, and each client has access to one of these subsets, picked at random. Each client has Python code for doing this training locally, which is then plugged to the Florida SDK by adding the lines in \ref{fig:sdk_python}.
We also need to send a script to the server with a recipe for aggregating model updates; in what follows, we have simply averaged them, as in the Federated Averaging scheme\cite{mcmahan2017communication}.

We use the AzureML simulator with 8 \texttt{Standard\_DS11\_v2} nodes containing 4 clients each, thus simulating 32 clients. At each of 10 iterations, each client accesses one of the 100 splits at random, and uses 20\% of the data in the split to update the model (approximately 67 samples). We use the trainer provided by the transformers library with its default optimizer (AdamW), a learning rate of $5 \cdot 10^{-4}$, and a batch size of 8. Note that the model we use has approximately 16Mb when compressed. The performance of the federated training is measured in terms of accuracy on the test set and duration of each iteration.

\begin{figure}[htp]
    \centering
    \includegraphics[width=0.32\textwidth]{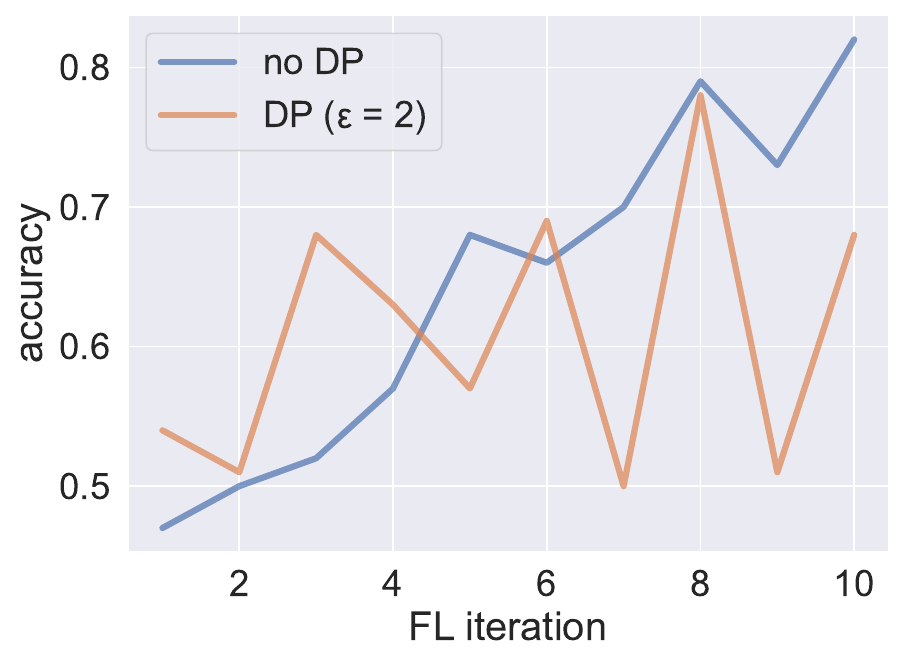}
    \includegraphics[width=0.32\textwidth]{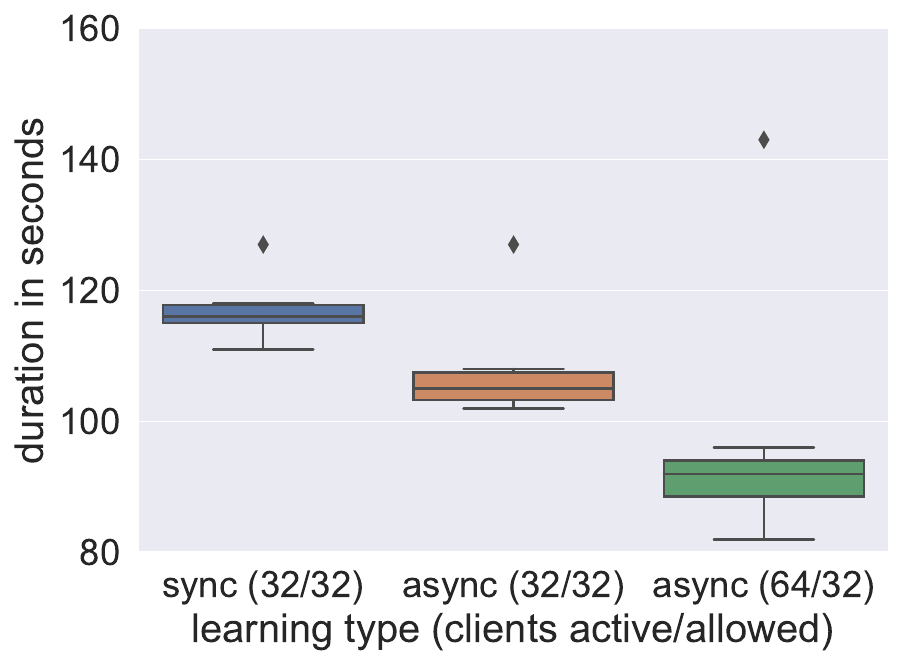}
    \includegraphics[width=0.32\textwidth]{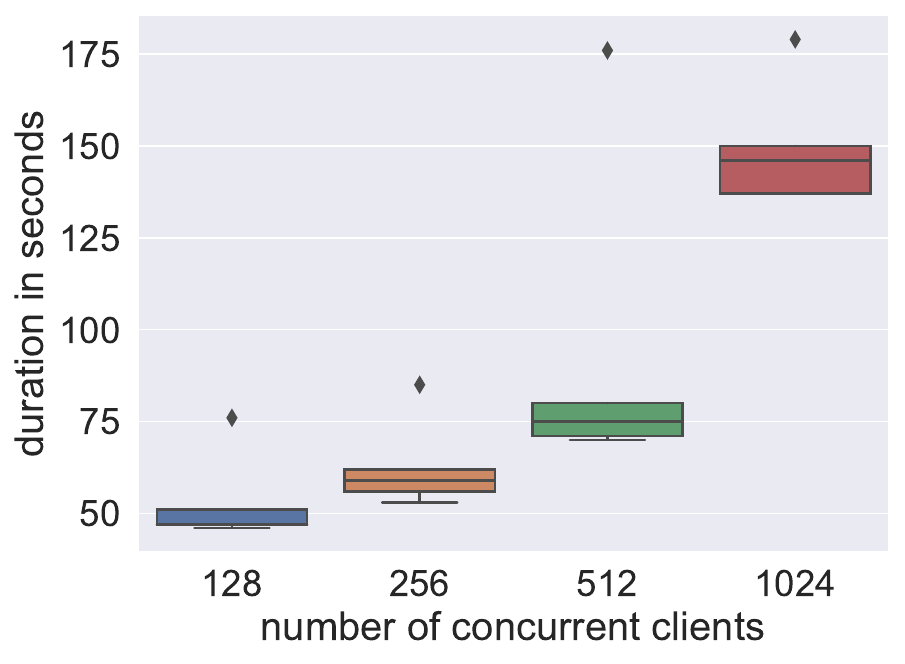}
    \caption{\emph{Left and Center:} Accuracy and duration of each iteration, for the spam classification task. When using differential privacy, accuracy slightly decreases, and convergence is affected. Duration of each iteration is shown to be smaller for asynchronous approaches. \emph{Right:} Scaling test showing the duration of each iteration for different numbers of concurrent clients, on a dummy task. Notice that the x-axis is not linear.}
    \label{fig:spam_results}
\end{figure}

Some variations of this experiment are attempted. First, we add user-level differential privacy (DP), by requesting the task to be created with local DP, a clipping norm of 0.5, and noise scale of 0.08; using the RDP accountant in Opacus, and considering there is a pool of 100 clients, we get a global $\varepsilon$ value of 2, with $\delta = 10^{-5}$. As we can see in Figure \ref{fig:spam_results} (left), this leads to a slight decrease in accuracy, as well as convergence issues, which are expected when using DP and can be mitigated by adjusting the hyper-parameters.

We can also change the type of learning to asynchronous. In that case, the ``iteration'' concept is lost; we instead use a buffer of size 32, i.e., the model is updated after every 32 pseudo-gradients received by the server, which is repeated for 10 times. As we can see in Figure \ref{fig:spam_results} (center), the average duration of each iteration decreases in that case, with similar accuracies obtained. We can also allow over-participation, by increasing the number of nodes to 16 instead of 8, which decreases the duration even further.

\subsection{Scaling Tests}

Finally, we run a dummy task on varying numbers of clients. We use the AzureML simulator with varying numbers of nodes containing 32 clients each. The task consists in having each client generating an all-ones array of size 5 and sending it to the server, which then aggregates all the arrays. As we can see in Figure~\ref{fig:spam_results} (right), we can get to the order of one thousand clients communicating concurrently with the server, while still having the iteration processed in a reasonable time. By spacing out the clients and increasing the iteration timeout, we can easily process hundreds of thousands of clients per iteration.

To date, Florida has been validated on production workloads of up to 70,000 connected devices, and our ongoing work is validating workloads of even greater capacity.

\section{Related Work}
Industrial applications of cross-device federated learning are largely represented by deployments from Google \cite{bonawitz2019towards} and Apple~\cite{paulik2021federated}.  Google has open-sourced TensorFlow Federated~\cite{Abadi+16} which provides basic building blocks for building FL training workflows, mainly from the perspective of machine learning, as opposed to devops.  Similarly, there are several platform vendors with an ML focus and use cases focused on cross-silo FL or collaborative learning, such as Flare from NVIDIA~\cite{roth2022nvidia}, PySyft~\cite{Ziller2021} and FedML~\cite{He+20}.  The Azure ML Platform also offers open source APIs for cross-silo FL based on the Azure platform~\cite{amlfl}. Several research groups have released simulation platforms for enabling FL research, including FedML~\cite{He+20}, Flower~\cite{Beutel+20}, and FLUTE~\cite{garcia2022flute}.

There is extensive literature exploring real-world considerations in FL, such as resilience to attack or bias due to diurnal rhythms~\cite{DBLP:journals/corr/abs-2108-10241,51174}. Generally these explorations are conducted off-line in simulation.  Google~\cite{bonawitz2019towards}, Facebook~\cite{DBLP:journals/corr/abs-2111-04877}, and Amazon~\cite{DBLP:journals/corr/abs-2202-03925} have all published work exploring real-world scaling and data bias considerations based on their experiences deploying real-world solutions.

\section{Discussion}
We have presented Florida, a cross-device orchestration platform for deploying FL solutions at scale, spanning heterogeneous device ecosystems.  Florida is designed to simplify FL deployment tasks by decoupling the training workflow from the complexities of service management and device orchestration.  Florida enables ML developers to focus on the local training and data life cycle and abstracts the necessary privacy and security tasks for securely aggregating the model. Florida has been validated to scale to tens of thousands of connected devices and we anticipate no serious difficulties scaling into the millions.

While Florida is fully-featured for many varieties of workloads, there are opportunities to add support for several enhancements: the aggregation approach, dependent as it is on secure aggregation, leaves limited room for novel aggregation approaches, such as clustered FL~\cite{DBLP:journals/corr/abs-1910-01991}, non-neural model architectures, or model architectures dependent on auxiliary metadata (for instance, updating and scaling min or max feature values).  Furthermore, secure aggregation may prohibit gradient compression techniques that become important for workflow scaling.  Our current work is exploring how these approaches can be more easily incorporated into secure aggregation schemes based on trusted enclaves, such as that proposed in ~\cite{DBLP:journals/corr/abs-2111-04877}.

Testing real-world FL workflows is also an open challenge. In most cases our testing was limited by our ability to scale a test harness of simulated clients. We are continuing efforts in this area, to enable testing and validation at scales sufficient to deploy novel solutions to millions of connected devices.

\bibliographystyle{unsrt}

\bibliography{references}

\end{document}